\pdfoutput=1

\documentclass[11pt]{article}

\usepackage[final]{acl}

\usepackage{times}
\usepackage{latexsym}

\usepackage[T1]{fontenc}

\usepackage[utf8]{inputenc}

\usepackage{microtype}
\microtypesetup{nopatch=footnote}

\usepackage{amsmath}
\usepackage{amssymb}
\usepackage{mathtools}

\usepackage{graphicx}
\usepackage{xspace}

\DeclareRobustCommand{\timepuzzleicon}{%
  \raisebox{-0.3\height}{\includegraphics[height=1.5em]{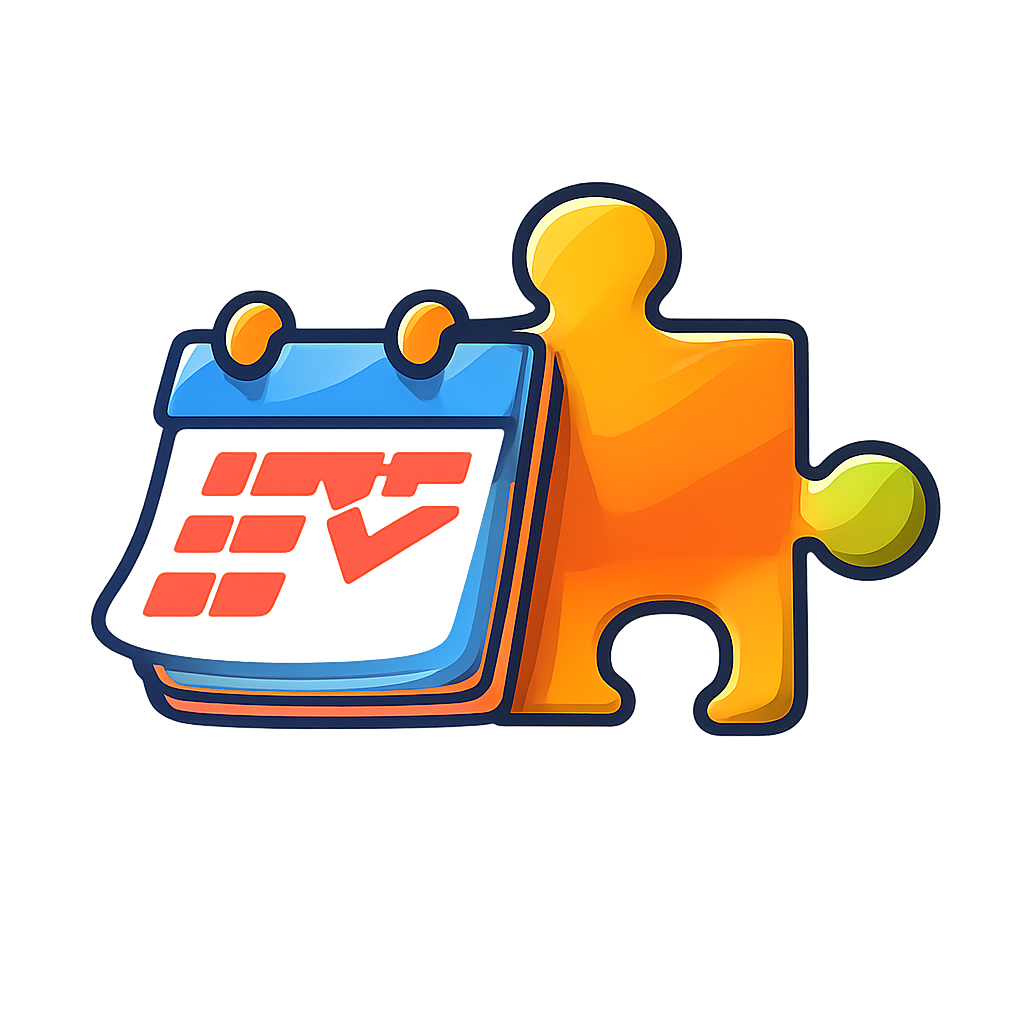}}%
}

\DeclareRobustCommand{\TimePuzzles}{\texorpdfstring{\timepuzzleicon\ Time Puzzles\xspace}{Time Puzzles}}

\usepackage{inconsolata}
\usepackage{todonotes}
\usepackage{booktabs}
\usepackage{tabularx}
\usepackage{array}
\usepackage{multirow}
\usepackage{siunitx}

\usepackage[ruled,vlined,linesnumbered]{algorithm2e}

\SetAlCapSty{myAlCapSty}
\SetAlFnt{\small}
\SetAlCapNameFnt{\small}
\SetAlgoNlRelativeSize{0}

\definecolor{jack}{rgb}{1.0, 0.6, 0.4}
\definecolor{zeyu}{rgb}{0.5, 0.2, 0.8}

\usepackage{listings}
\usepackage{xcolor}
\usepackage{setspace}
\definecolor{aclblue}{RGB}{30, 80, 120}
\definecolor{promptgray}{RGB}{248, 248, 248}
\definecolor{emgreen}{RGB}{0,120,60}     
\definecolor{tokred}{RGB}{160,60,60}

\lstdefinestyle{promptstyle}{
  basicstyle=\ttfamily\footnotesize,
  columns=fullflexible,
  breaklines=true,
  breakindent=0pt,
  breakautoindent=false,
  frame=single,
  framerule=0.5pt, 
  rulecolor=\color{aclblue}, 
  backgroundcolor=\color{promptgray}, 
  xleftmargin=1em,
  xrightmargin=1em,
  aboveskip=1em,
  belowskip=1em,
  showstringspaces=false,
  keepspaces=true
}

\title{Measuring Iterative Temporal Reasoning with \TimePuzzles}

\author{Zhengxiang Wang\thanks{Equal contributions. ZW conceived the research idea. ZD designed and implemented the algorithm for dataset generation. ZW designed and executed the LLM experiments. Both authors contributed significantly to drafting and revising the manuscript. ZW revised the second version of this draft.} \\
  Department of Linguistics \& IACS \\
  Stony Brook University \\
  \texttt{zhengxiang.wang@stonybrook.edu} \\\And
  Zeyu Dong\footnotemark[1] \\
  Department of Applied Math and Statistics \\
  Stony Brook University \\
  \texttt{zeyu.dong@stonybrook.edu} \\}

\begin{document}
\maketitle
\begin{abstract}
Tool use, such as web search, has become a standard capability even in freely available large language models (LLMs). However, existing benchmarks evaluate temporal reasoning mainly in static, non-tool-using settings, which poorly reflect how LLMs perform temporal reasoning in practice. We introduce \TimePuzzles, a constraint-based date inference task for evaluating \emph{iterative temporal reasoning with tools}. Each puzzle combines factual temporal anchors with (cross-cultural) calendar relations and may admit one or multiple valid dates. The puzzles are algorithmically generated, enabling controlled and continual evaluation. Across 13 LLMs, even the best model (GPT-5) achieves only 55.3\% accuracy without tools, despite using easily searchable facts. While web search improves performance, models perform substantially better when constraints are rewritten with explicit dates, removing the need for factual lookup. These results reveal a gap in reliable tool use for iterative temporal reasoning.


\end{abstract}



\section{Introduction}
\label{sec:intro}

Modern LLMs are increasingly deployed with external tools, such as web search and code interpreter, even in free customer-facing settings \citep{openrouter2025stateofai}. These capabilities allow models to retrieve factual information and verify intermediate reasoning steps during inference. However, existing temporal reasoning benchmarks largely evaluate models in \emph{static, single-shot} settings, which do not reflect the iterative, tool-augmented workflows increasingly used in practice. 

Prior work on temporal reasoning has focused on static, single-shot settings \citep{ning-etal-2020-torque,zhou-etal-2021-temporal,zhou-etal-2019-going,qin-etal-2021-timedial}. While recent benchmarks extend evaluation to LLMs and situated contexts \citep{tan-etal-2023-towards,chu-etal-2024-timebench,wang-zhao-2024-tram,wei-etal-2025-time-benchmark,chen-etal-2021-time-sensitive-questions}, they still do not explicitly evaluate tool-augmented, iterative temporal reasoning



\begin{figure}
    \centering
    \includegraphics[width=0.95\linewidth]{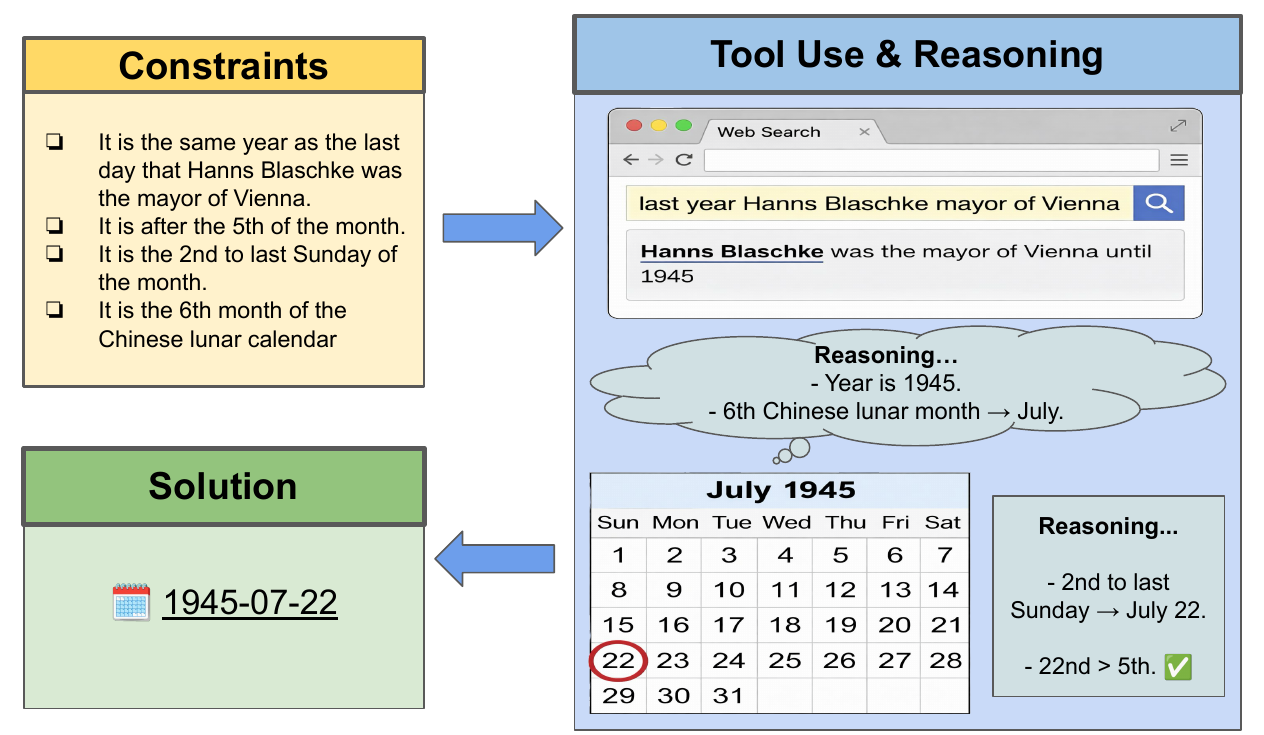}
    \caption{\TimePuzzles: a simple date inference task requiring iterative tool-aided temporal reasoning.}
    \label{fig:timePuzzlesIllustration}
\end{figure}

To address this gap, we propose \TimePuzzles, a family of simple, puzzle-like constraint-based date inference tasks designed to evaluate \textit{precise, iterative} temporal reasoning with tools. As illustrated in Figure~\ref{fig:timePuzzlesIllustration}, each puzzle presents natural-language temporal constraints that combine factual temporal anchors with (cross-cultural) calendar-based relations, and asks the model to infer the date(s) that jointly satisfy all constraints. Solving a puzzle requires \emph{iteratively proposing, refining, and verifying candidate dates} using external tools (e.g., calendars or web search), emphasizing constraint-based reasoning rather than rote memorization or single-shot recall. This formulation reflects real-world settings such as scheduling, planning, and historical analysis, where systems must integrate factual information with calendar structure to produce temporally consistent decisions.

We evaluate 13 diverse LLMs on \TimePuzzles under both tool-less and tool-augmented settings. Overall performance remains limited: even the best model (GPT-5) reaches only 55.3\% accuracy without tools, despite the use of easily searchable factual anchors. While web search consistently improves results, all models perform better when the same puzzles are rewritten to replace factual anchors with explicit dates. This gap highlights a key weakness in integrating factual lookup with constraint-based, multi-step temporal reasoning.






\section{\TimePuzzles}\label{sec:task}


This section defines the \TimePuzzles task and describes its algorithmic generation process.

\subsection{Task Formulation}\label{subsec:task_formulation}

\TimePuzzles is a constraint-based date inference task: given a set of temporal constraints, the goal is to identify the date(s) that jointly satisfy all constraints. Formally, let $\mathcal{D}$ denote the set of all possible Gregorian dates, and let $\mathcal{C}$ be an oracle function that maps a natural-language temporal constraint $t$ to the set of dates for which $t$ holds, i.e., $\mathcal{C}(t)\subseteq\mathcal{D}$. As illustrated in Figure~\ref{fig:timePuzzlesIllustration}, each puzzle consists of $N$ natural-language constraints $F=\{t_1,t_2,\dots,t_N\}$. The answer set $\mathcal{A}$ contains all dates that satisfy every constraint:

\begin{equation}
    \mathcal{A} = \bigcap_{i=1}^{N} \mathcal{C}(t_i)
\end{equation}

We restrict our study to puzzles with non-empty answer sets ($|\mathcal{A}| \ge 1$) to focus on LLMs' ability to identify valid solution dates via iterative temporal reasoning, and cap $|\mathcal{A}|$ at 6 for exploration.

\subsection{Data Generation}\label{subsec:data_generation}

\paragraph{Input Constraints} We create a simple yet diverse taxonomy of temporal constraint types (Table~\ref{tab:fact_taxonomy} in Appendix~\ref{app:algorithms}). The taxonomy spans both factual anchors (e.g., historical events, presidents in office, zodiac years) and calendar-structural constraints (e.g., months, seasons, weekdays, or day-of-month relations), each restricting candidate dates at the year, month, or day level.


Because \TimePuzzles is designed to probe tool-augmented iterative reasoning, we append to each puzzle a randomly sampled, trivial historical event from a curated list of 50 that \emph{potentially} require web search. All events occur no later than 2023 and are \emph{easily searchable}, as manually verified during curation (Appendix~\ref{app:factCollection}). We leave harder-to-find facts for future stress-testing.

This paper considers only English as a proof of concept, but \TimePuzzles can be readily adapted to other languages. Figure~\ref{fig:timePuzzlesIllustration} shows an example puzzle with five constraints.


\paragraph{Puzzle Generation} Puzzles are generated algorithmically through randomized constraint sampling. Each puzzle combines one of the 50 real-world historical events we manually curated  with additional temporal constraints from the taxonomy (e.g., calendar relations). We retain combinations that yield puzzles with a desired number of valid solution dates. We open-source our code and generated datasets at \url{https://github.com/jaaack-wang/Time-Puzzles}.

\paragraph{Resulting Datasets} We construct two conceptually equivalent datasets for controlled evaluation. The default dataset uses implicit constraints (e.g., an event) to target tool-augmented iterative temporal reasoning. The paired dataset rewrites each implicit constraint by spelling out its relevant dates explicitly, thereby removing the need of factual lookup. Each dataset contains 600 puzzles, with solution set sizes evenly distributed from 1 to 6. 

We select 100 puzzles per solution size due to cost constraints, as unrestricted web search is prohibitively expensive, given our limited budget. In addition, experiments on a separate set of 600 puzzles yield similar results, supporting the suitability of the current dataset size (see Section~\ref{sec:analysis}).

\paragraph{Quality Control} Because all constraints are independently verifiable, the resulting puzzles are both scalable and high quality. We randomly inspected 30 puzzles, where one author independently attempted to solve them using a web browser and verified the correctness of the generated puzzles. This manual inspection also yielded an estimated human baseline accuracy of 80\%.

\begin{figure*}[hbt!]
    \centering
    \small
    \includegraphics[width=1\linewidth]{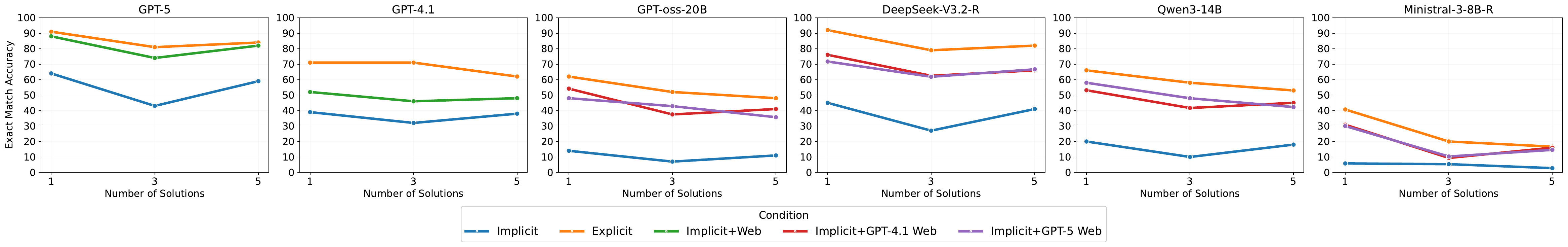}

    \caption{Average exact match accuracy across solution counts, with and without web search (only on solution counts 1, 3, 5). GPT-4.1/5 run live web search; for open-weight models we re-use the same \textbf{cited} GPT web results.}

    \label{fig:em_across_solution_counts}
\end{figure*}

\section{Experiments\label{sec:experiments}}

This section first outlines the overall experimental setup and then reports results on \TimePuzzles under both tool-free and tool-augmented settings.

\subsection{Experimental Setup}

\paragraph{Models} We evaluate 13 diverse LLMs from four model families (Table~\ref{tab:tool_less_overall_results}) released in 2025. These include four proprietary GPT models (GPT-5 and GPT-4.1 series) \cite{openai_gpt41_2025} and nine open-weight models: GPT-oss-20B \cite{openai2025gptoss120bgptoss20bmodel} plus instruction/reasoning variants from DeepSeek-V3.2 \cite{deepseekai2025deepseekv32pushingfrontieropen}, Qwen3 \cite{yang2025qwen3technicalreport}, and Ministral-3-8B \cite{mistral_mistral3_2025}. See Table~\ref{tab:model_details} in Appendix~\ref{app:models} for further details.

\paragraph{Prompting} All models are prompted using zero-shot chain-of-thought (CoT) prompting \cite{zero-shot-cot} to encourage step-by-step reasoning. The prompt template is provided in Appendix~\ref{app:prompts}.

\begin{table}
\centering
\scriptsize
\setlength{\tabcolsep}{3.5pt}
\begin{tabular}{lrrrr}
\toprule
Model & EM (\%) & JI (\%) & F1 (\%) & \# Tks \\
\midrule

GPT-5 & 55.33 & 58.62 & 59.52 & 4126 \\
GPT-5-nano & 15.83 & 19.37 & 20.31 & 5224 \\ \midrule 
GPT-4.1 & 35.17 & 39.13 & 40.14 & 615 \\
GPT-4.1-nano & 9.33 & 13.53 & 14.84 & 1098 \\ \midrule 

GPT-oss-20B & 10.83 & 13.51 & 14.32 & 5704 \\ \midrule

DeepSeek-V3.2-R & 36.67 & 39.33 & 40.08 & 7862 \\
DeepSeek-V3.2 & 30.50 & 32.89 & 33.51 & 737 \\ \midrule

Qwen3-4B-Inst & 3.00 & 3.89 & 4.14 & 2408 \\
Qwen3-4B-Think & 5.67 & 6.61 & 6.89 & 9701 \\
Qwen3-8B & 10.83 & 13.45 & 14.26 & 7341 \\
Qwen3-14B & 13.17 & 16.32 & 17.36 & 4733 \\ \midrule

Ministral-3-8B-Inst & 3.50 & 5.38 & 6.04 & 1797 \\
Ministral-3-8B-R & 4.27 & 4.81 & 5.04 & 2058 \\

\bottomrule
\end{tabular}

\caption{Exact-match accuracy (EM), Jaccard index (JI), F1 score (F1), and output tokens (Tks), averaged over six solution counts, for the 600 generated puzzles with implicit (default) constraints in the tool-less setting.}
\label{tab:tool_less_overall_results}

\end{table}

\paragraph{Metrics} Let $Y_i$ denote the gold solution set and $\hat{Y}_i$ the predicted set for instance $i$. We compute exact match (EM), F1 (F1), and Jaccard index (JI):

\begin{equation}
EM_i=\mathbb{I}[Y_i=\hat Y_i].
\end{equation}

\begin{equation}
P_i=\tfrac{|Y_i\cap\hat Y_i|}{|\hat Y_i|},\;
R_i=\tfrac{|Y_i\cap\hat Y_i|}{|Y_i|},\;
F1_i=\tfrac{2P_iR_i}{P_i+R_i}.
\end{equation}

\begin{equation}
JI_i=
\begin{cases}
1 & Y_i=\hat Y_i=\emptyset \\
\tfrac{|Y_i\cap\hat Y_i|}{|Y_i\cup\hat Y_i|} & \text{otherwise}.
\end{cases}
\end{equation}

Given our emphasis on \emph{precise} iterative reasoning, we treat EM as the primary metric.


\paragraph{Tools}
We consider the following tool-augmented settings: web search, Code Interpreter (+CI), and their combinations. Due to cost constraints, these experiments are conducted on puzzles with solution counts 1, 3, and 5 and with selected models. 

More specifically, for web search, we test six best-performing LLMs across different model families (GPT-5, GPT-4.1, GPT-oss-20B, DeepSeek-V3.2-R, Qwen3-14B, and Ministral-3-8B-R), whereas for +CI, we only evaluate GPT-5 and GPT-4.1 both with and without web search. We use off-the-shelf tools via OpenAI Response API and we include GPT-5 and GPT-4.1 across all tool-augmented settings to represent tool-using capablities by a reasoning model (GPT-5) as well as an instruction model (GPT-4.1). Note that GPT-5 and GPT-4.1 are run with live web search and their retrieved results are reused as fixed context for the open-weight models. 




\subsection{Results}


\paragraph{Despite its simplicity, \TimePuzzles is challenging for tool-less LLMs.} Our dataset is a \emph{proof of concept} built from easy-to-search facts that predates each model’s training cutoff or release date. Yet even GPT-5 achieves only 55.3\% EM on the dataset, with all other models below 37\% EM (Table~\ref{tab:tool_less_overall_results}). JI and F1 yield slightly higher scores, but the overall performance patterns remain consistent, with a Spearman correlation of $r = 0.93$. Moreover, the three metrics agree in 88.6\% of cases, indicating that LLMs typically either solve a puzzle completely or fail to solve it at all.

\begin{figure}
    \centering
    \small
    \includegraphics[width=0.95\linewidth]{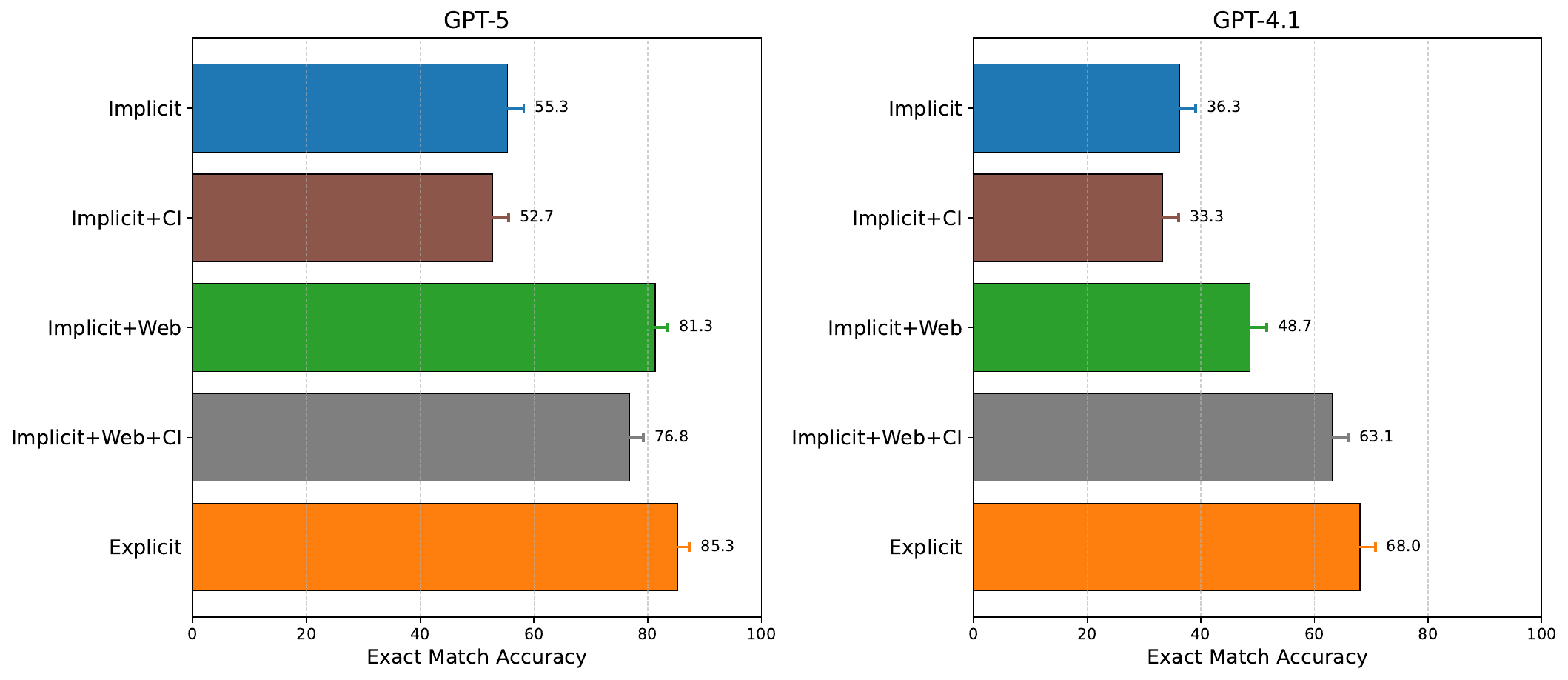}
    \caption{Average EM accuracy (\%) for GPT-5 (left) and GPT-4.1 (right) under different conditions across three solution counts (1, 3, and 5). Numbers annotated on bars are mean values, beside error bars. \texttt{+Web} enables web search. \texttt{+CI} enables Code Interpreter.}
    \label{fig:code_interpreter_conditions_comparison}
\end{figure}

\paragraph{Tool use, particularly web search, can greatly improve model performance, but a clear reasoning gap remain.} Figure~\ref{fig:em_across_solution_counts} shows that web search consistently improves performance across different solution counts, but models nearly always perform the best when the implicit constraints in \TimePuzzles are rewritten with explicit dates to remove the need for factual lookup. Similarly, Figure~\ref{fig:code_interpreter_conditions_comparison} shows that enabling Code Interpreter (+CI) does not close the gap either: +CI degrades GPT-5 performance on implicit constraints, both with and without web search. While GPT-4.1 benefits from +CI when combined with web search, the performance is below that of explicit constraints.

\section{Analysis\label{sec:analysis}}

Section~\ref{sec:experiments} shows that \TimePuzzles provides a challenging and cost-effective benchmark for iterative temporal reasoning. This section further demonstrates that as a benchmark, \TimePuzzles is both discriminative and reliable. 

\paragraph{\TimePuzzles effectively distinguishes nuanced iterative temporal reasoning capabilities.} Table~\ref{tab:tool_less_overall_results} clearly shows that larger models consistently outperform their smaller counterparts (e.g., GPT-X vs.\ GPT-X-nano; Qwen3 variants), and reasoning models uniformly surpass instruction-tuned ones (e.g., DeepSeek/Ministral pairs). However, scale alone is insufficient: despite its smaller size, Qwen3-4B-Think substantially outperforms Ministral-3-8B-R, and GPT-oss-20B similarly surpasses both while using only 3.6B active parameters. Moreover, for reasoning models, higher performance typically uses less output tokens, suggesting that success on \TimePuzzles hinges on sustained, structured reasoning rather than exhaustive generation (e.g., Qwen3 models in Table~\ref{tab:tool_less_overall_results}).

\paragraph{Our findings are robust to prompt variations and puzzle sets.}
Table~\ref{tab:follow_up_results} shows that inserting a detailed step-by-step reasoning instruction into the default prompt (see Appendix~\ref{app:prompts}) decreases performance, particularly for larger models. Furthermore, repeating the experiments from Section~\ref{sec:experiments} on 600 newly generated puzzles yields negligibly different exact-match (EM) scores, validating the suitability of our evaluation dataset size.

\begin{table}[t]
\centering
\scriptsize
\begin{tabular}{llcccc}
\toprule
\textbf{Type} & \textbf{Setting} & \textbf{G4.1-Nano} & \textbf{G5-Nano} & \textbf{G4.1} & \textbf{G5} \\
\midrule

\multirow{3}{*}{Implicit}
 & Default      & 9.3  & 15.8 & 35.2 & 55.3 \\
 & +Reason Inst & 8.8  & 14.2 & 25.3 & 47.3 \\
 & New Runs     & 11.0 & 16.8 & 29.8 & 50.1 \\

\midrule

 \multirow{3}{*}{Explicit}
 & Default      & 41.8 & 60.2 & 63.0 & 84.7 \\
 & +Reason Inst & 35.5 & 55.5 & 53.5 & 77.7 \\
 & New Runs     & 42.7 & 60.2 & 63.7 & 80.5 \\

\bottomrule
\end{tabular}
\caption{EM accuracy (\%) averaged over six solution counts under different settings for both explicit and implicit constraint across the four GPT models. Default setting is based on Section~\ref{sec:experiments}. ``+Reason Inst'' enhances the default setting with detailed step-by-step reasoning instruction. ``New Runs'' reruns the default setting with completely new puzzles generated from scratch.}
\label{tab:follow_up_results}
\end{table}

\section{Related Work}

Temporal reasoning has long been studied in NLP, including event ordering, implicit event inference, and temporal commonsense reasoning \citep{ning-etal-2020-torque,zhou-etal-2021-temporal,zhou-etal-2019-going,qin-etal-2021-timedial}. More recent work focuses on diagnosing temporal reasoning in LLMs through broader benchmarks and systematic analyses across diverse task formats \citep{tan-etal-2023-towards,chu-etal-2024-timebench,wang-zhao-2024-tram,wei-etal-2025-time-benchmark}, as well as targeted probes of specific temporal phenomena and synthetic constructions \citep{islakoglu-kalo-2025-chronosense,wei-etal-2023-menatqa,fatemi-etal-2025-test-of-time,bhatia-etal-2025-datelogicqa}.

A complementary line of work studies temporal reasoning in situated and dynamic contexts \citep{chen-etal-2021-time-sensitive-questions,zhang-choi-2021-situatedqa,kasai-etal-2023-realtimeqa}. Other studies explore more agentic settings that incorporate memory, external computation, or multimodal inputs \citep{ge-etal-2025-tremu,saxena-etal-2025-lost-in-time}. Overall, existing benchmarks largely evaluate temporal reasoning in static or retrieval-oriented settings. In contrast, \TimePuzzles focus on constraint-based date inference that benefits from explicit tool use and requires iterative reasoning, providing a complementary evaluation of tool-augmented, agentic temporal reasoning.

\section{Conclusion}

We propose \TimePuzzles, a constraint-based date inference task that targets iterative, tool-augmented temporal reasoning. Although puzzles are synthetically generated and easy to verify, they are challenging for tool-less LLMs. They also expose persistent failures of current instruction- and reasoning-tuned LLMs to reliably resolve implicit temporal constraints, even with tool access. Overall, \TimePuzzles offers a simple, cost-effective, and discriminative diagnostic for tool-augmented iterative temporal reasoning and can be systematically extended to support more challenging evaluations or additional languages (e.g., via translation).

\section*{Acknowledgement}

Zhengxiang Wang is grateful for support from the Institute for Advanced Computational Science (IACS) at Stony Brook University. We also thank the three anonymous reviewers for their valuable feedback.

\section*{Limitations} 

\paragraph{Synthetic Dataset Construction}
\TimePuzzles is enabled by a controllable generation algorithm, but the resulting constraints are not guaranteed to match real-world temporal reasoning in full complexity. In particular, the naturalness of the constraints can vary, and template-based composition may miss ambiguity, conflicting evidence, or underspecified conditions that arise in practice.

\paragraph{Tool and Model Scope}
Our experiments do not exhaust the space of models, tools, or agentic workflows. We evaluate off-the-shelf LLMs and off-the-shelf tools, and we do not attempt to build sophisticated tool-use policies to solve Time Puzzles, given the scope of the study. As a result, our results should be interpreted as a diagnostic of current out-of-the-box capabilities rather than an upper bound.

\paragraph{Solution-Less Puzzles} We restrict evaluation to puzzles with at least one valid solution in order to focus on LLMs’ ability to identify correct dates via iterative temporal reasoning. Although our generation algorithm can produce puzzles with zero solutions, such instances often contain trivial self-contradictions (e.g., incompatible year, month, or day constraints) that require little reasoning and are easily solved by smaller models. Designing non-trivial solution-less puzzles that demand substantive iterative reasoning is left to future work.

\paragraph{Language Diversity} Our work currently only considers English language, due to budget constaints. Future studies can extend our study by adapting our evaluation framework to other languages. This can be easily done by, for example, converting the puzzle template into other languages.

\section*{Ethical Considerations}

\paragraph{Data and privacy.}
We do not collect human subjects data. Time Puzzles instances are synthetically generated, and our factual anchors are based on public historical information, with no personally identifiable information.

\bibliography{custom}

\clearpage

\appendix

\section{Data Generation \label{app:data_generation}}

\subsection{Factual Anchor Collection\label{app:factCollection}}

\paragraph{Collection Procedure} We manually curate 50 easy-to-search factual instances based on TEMPREASON \cite{tan-etal-2023-towards}, a time-sensitive question answering dataset designed to evaluate temporal reasoning over time--event and event--event relations. Our collection procedure is as follows:

\begin{itemize}
    \item We randomly sample 400 questions involving time--event and event--event relations from TEMPREASON. Each instance consists of a factual question (e.g., ``Which position did Siim Kallas hold in September 1999?'') and its corresponding answer (e.g., ``Minister of Finance'').
    
    \item We prompt \texttt{gpt-oss-20b} to answer the sampled questions and find that it could barely answer these questions. 
    
    \item For each retained instance, we reframe the question--answer pair into a declarative statement describing an event, with the temporal information removed. We then use web search to recover the corresponding time point or time range during which the event occurred. When necessary, we further reframe the instance to query a closely related event with an unambiguous temporal answer. We stop once 50 valid instances are collected. \textbf{All temporal answers are obtained from top-ranked Google search results and manually verified by the authors.}
    
    \item We use browser-based ChatGPT-5.2-Auto and Gemini 3 Pro with web search enabled to independently fact-check the complete fact table (provided via screenshots). Minor typographical errors are corrected, and one ambiguous instance is replaced. After revision, both models identify no remaining issues.
\end{itemize}

\paragraph{Fact-Checking Prompt Instruction} See below.

\begin{lstlisting}[style=promptstyle]
You are given a table containing 50 historical events. For each event, perform a web search to verify whether the provided time information is accurate.

Column meanings:

- start: The year (or date) when the event first began.

- end: The year (or date) when the event ended.

- multi_year_spans: The range of years during which the event took place.

Task instructions:

- Verify whether the provided time information for each event is correct or incorrect based on reliable sources.

- The time may be given at different levels of precision (e.g., year-only vs. full date).

- Judge correctness only at the precision provided.

- Do not mark a time as incorrect solely because it omits a more specific month or day.

Output requirements (one entry per event):
For each event, report:

- The row index

- The event name

- Whether the provided time is correct or incorrect

- A brief explanation supporting your judgment
\end{lstlisting}

\paragraph{Including Well-Known Facts} To balance the corpus, we additionally include 20 widely known factual instances. These cover the presidencies of recent U.S.\ presidents (since 1945) and the lifespans of well-known public figures (e.g., Kobe Bryant, Albert Einstein).

\subsection{Algorithm Implementation Details}\label{app:algorithms}
We generate each puzzle by sampling a hidden \emph{seed date} and selecting $N$ natural-language constraints whose conjunction yields a user-defined number of solutions. To ensure diversity and computational efficiency, our pipeline utilizes a comprehensive taxonomy of temporal facts and sorts constraints by \textit{Information Gain} (IG) during verification.

\paragraph{Fact Taxonomy and Constraint Levels}

To ensure diversity, we define a set of fact categories $\mathcal{K}$. Each category represents a specific type of temporal knowledge (e.g., calendar structure, historical events, astronomical cycles). Furthermore, we assign a \textit{Constraint Level} $L(t) \in \{\textsc{Year}, \textsc{Month}, \textsc{Day}\}$ to each fact type, indicating the granularity of the information provided. Table~\ref{tab:fact_taxonomy} summarizes the fact categories used in \TimePuzzles.

\begin{table*}[ht]
\centering
\small
\begin{tabular}{l p{6cm} c}
\toprule
\textbf{Fact Name} & \textbf{Description} & \textbf{Level} \\
\midrule
\texttt{ExplicitYearFact} & Explicitly states the year (e.g., ``The year is 2025''). & \textsc{Year} \\
\texttt{DecadeFact} & Specifies the decade (e.g., ``The 1990s''). & \textsc{Year} \\
\texttt{LeapYearFact} & States that the year is a leap year. & \textsc{Year} \\
\texttt{ChineseZodiacFact} & Specifies the Chinese Zodiac animal. & \textsc{Year} \\
\texttt{PersonAliveFact} & States a famous person (e.g., Steve Jobs) was alive. & \textsc{Year} \\
\texttt{USPresidentFact} & States a specific US President was in office. & \textsc{Year} \\
\texttt{EventFact} & States if it is an Olympic or World Cup year. & \textsc{Year} \\
\midrule
\texttt{MonthFact} & Specifies the month of the year. & \textsc{Month} \\
\texttt{SeasonFact} & Specifies the season (e.g., Winter) based on month. & \textsc{Month} \\
\texttt{LunarMonthFact} & Specifies the month in the Chinese lunar calendar. & \textsc{Month} \\
\midrule
\texttt{WeekdayInMonthFact} & Specifies the Nth occurrence of a weekday (e.g., ``2nd Friday''). & \textsc{Day} \\
\texttt{DayOfMonthFact} & Specifies the exact day or if it is the first/last day. & \textsc{Day} \\
\texttt{WeekdayFact} & Specifies the day of the week (e.g., ``It is Monday''). & \textsc{Day} \\
\texttt{MultiWeekdayFact} & Specifies a set of possible weekdays (e.g., ··Mon or Tue''). & \textsc{Day} \\
\texttt{DayOfMonthRangeFact} & Specifies if the day is before/after a specific day. & \textsc{Day} \\
\midrule
\texttt{KnowledgeBaseEventFact} & Relates the date to a historical event (same day/month/year), such as the 50 facts collected in Appendix~\ref{app:factCollection}. & \textsc{Various} \\
\bottomrule
\end{tabular}
\caption{Taxonomy of temporal facts used in generation, categorized by constraint granularity level.}
\label{tab:fact_taxonomy}
\end{table*}

\paragraph{Entropy and Information Gain}

To generate efficient solvers and measure the utility of each clue, we define the IG for a fact $t$. Let the entropy of the date space given a set of constraints be the logarithm of the size of the valid date set. The information gain of a fact $t$ is the reduction in entropy it provides:

\begin{equation}
IG(t) = \log_2(|\mathcal{D}|) - \log_2(|\mathcal{C}(t)|)
\end{equation}

Facts with higher IG narrow the search space more aggressively. For example, an exact date has maximal IG, whereas ``It is a Monday'' has low IG.

\paragraph{Algorithm Details}
We provide the specific procedural logic for generating diverse fact sets and solving the puzzles, as referenced in the main methodology. The fact generation process, detailed in Algorithm~\ref{alg:generate_fact_detail}, ensures diversity by selecting facts from different categories and balanced constraint levels. The solution verification process, detailed in Algorithm~\ref{alg:solve_puzzle_detail}, optimizes performance by first filtering with facts with high IG.

\begin{algorithm}[ht]
\SetAlgoLined
\KwIn{Anchor $t_{kb}$, Seed $d_{seed}$, Count $N$, Universal Date Set $\mathcal{D}$}
\KwOut{Set of facts $F$}
  $F \leftarrow \{t_{kb}\}$\;
  $UsedCats \leftarrow \{Cat(t_{kb})\}$\;
  
  \tcp{Iteratively generate N-1 distinct facts}
  \For{$i \leftarrow 1$ \KwTo $N-1$}{
    \tcp{Ensure balanced levels}
    $TargetLevel \leftarrow \text{DetermineLevel}(i, N)$\;
    $CandidateCats \leftarrow \{k \in \mathcal{K} \mid k \notin UsedCats, Level(k) == TargetLevel\}$\;
    $k_{new} \leftarrow \text{RandomSelect}(CandidateCats)$\;
    $t_{new} \leftarrow \text{InstantiateFact}(k_{new}, d_{seed})$\;
    $F \leftarrow F \cup \{t_{new}\}$\;
    $UsedCats \leftarrow UsedCats \cup \{k_{new}\}$\;
  }
  \Return $F$\;
 \caption{\textsc{GenerateFact} Procedure}
 \label{alg:generate_fact_detail}
\end{algorithm}

\vspace{1em}

\begin{algorithm}[ht]
\SetAlgoLined
\KwIn{Set of facts $F$, Universal Date Set $\mathcal{D}$}
\KwOut{Solution set $\mathcal{A}$}
  \tcp{Sort facts to optimize intersection}
  Compute $IG(t)$ for all $t \in F$\;
  $F_{sorted} \leftarrow \text{SortDescending}(F, \text{key}=IG)$\;
  
  $\mathcal{A} \leftarrow \mathcal{D}$\;
  \For{$t \in F_{sorted}$}{
    $\mathcal{A} \leftarrow \mathcal{A} \cap \mathcal{C}(t)$\;
  }
  \Return $\mathcal{A}$\;
 \caption{\textsc{SolvePuzzle} Procedure}
 \label{alg:solve_puzzle_detail}
\end{algorithm}

\begin{table*}[t]
\centering
\scriptsize
\setlength{\tabcolsep}{5pt}
\renewcommand{\arraystretch}{1.15}
\begin{tabularx}{\textwidth}{@{}p{4.5cm} p{4.9cm} p{2.1cm} p{1.7cm} X@{}}
\toprule
\textbf{Model} & \textbf{Variant used (snapshot / HF repo)} & \textbf{Params} & \textbf{Release date} & \textbf{Knowledge cutoff} \\
\midrule

GPT-5 \cite{openai_gpt5} & \texttt{gpt-5-2025-08-07} & Not disclosed & 2025-08-07 & 2024-09-30 \\
GPT-5-nano \cite{openai_gpt5} & \texttt{gpt-5-nano-2025-08-07} & Not disclosed & 2025-08-07 & 2024-05-31 \\
GPT-4.1 \cite{openai_gpt41_2025} & \texttt{gpt-4.1-2025-04-14} & Not disclosed & 2025-04-14 & 2024-06-01 \\
GPT-4.1-nano \cite{openai_gpt41_2025} & \texttt{gpt-4.1-nano-2025-04-14} & Not disclosed & 2025-04-14 & 2024-06-01 \\
\midrule

GPT-oss-20B \cite{openai2025gptoss120bgptoss20bmodel} & \texttt{openai/gpt-oss-20b} & 20.9B (3.6B active) & 2025-08-05 & 2024-06 \\
\midrule

DeepSeek-V3.2-R \cite{deepseekai2025deepseekv32pushingfrontieropen} & \texttt{deepseek-reasoner} & 685B & 2025-12-01 & Not disclosed \\
DeepSeek-V3.2 \cite{deepseekai2025deepseekv32pushingfrontieropen}  & \texttt{deepseek-chat} & 685B & 2025-12-01 & Not disclosed \\
\midrule

Qwen3-4B-Inst \cite{yang2025qwen3technicalreport} & \texttt{Qwen/Qwen3-4B-Instruct-2507} & 4B & 2025-08-06 & Not disclosed \\
Qwen3-4B-Think \cite{yang2025qwen3technicalreport} & \texttt{Qwen/Qwen3-4B-Thinking-2507} & 4B & 2025-08-06 & Not disclosed \\
Qwen3-8B \cite{yang2025qwen3technicalreport} & \texttt{Qwen/Qwen3-8B} & 8B & 2025-04-29 & Not disclosed \\
Qwen3-14B \cite{yang2025qwen3technicalreport} & \texttt{Qwen/Qwen3-14B} & 14B & 2025-04-29 & Not disclosed \\
\midrule

Ministral-3-8B-Inst \cite{mistral_mistral3_2025} & \texttt{mistralai/Ministral-3-8B-Instruct-2512} & 8B & 2025-12-02 & Not disclosed \\
Ministral-3-8B-R  \cite{mistral_mistral3_2025} & \texttt{mistralai/Ministral-3-8B-Reasoning-2512} & 8B & 2025-12-02 & Not disclosed \\
\bottomrule
\end{tabularx}

\caption{Model variants and specifications used in our experiments. We identify OpenAI models by their dated snapshot IDs and open-weight models by their Hugging Face repository IDs. Release dates reflect snapshot timestamps or official provider documentation. Note that precise release and knowledge cutoff dates may vary slightly across sources.}
\label{tab:model_details}
\end{table*}

\section{Models \label{app:models}} 

\paragraph{Inference Details} We evaluate four OpenAI GPT-family models via the OpenAI API (GPT-5, GPT-5-nano, GPT-4.1, and GPT-4.1-nano), using the dated snapshot identifiers shown in Table~\ref{tab:model_details}. We also evaluate two DeepSeek-V3.2 endpoints via the DeepSeek API (\texttt{deepseek-chat} and \texttt{deepseek-reasoner}; accessed in December 2025). For open-weight models, we run local inference with vLLM~\cite{kwon2023efficient}. We use default generation configuration for all models.

\paragraph{Model Specification} Table~\ref{tab:model_details} lists the exact model variants (API snapshot IDs or Hugging Face repo IDs), parameter sizes, release dates, and (when disclosed) knowledge cutoffs.

\paragraph{Web Search Implementation}
We enable web search for GPT-4.1 and GPT-5 via the official Responses API by providing the web search tool at inference time and setting tool choice as \emph{required}. From each model's output JSON, we extract the URLs cited in its response and scrape the corresponding webpage contents to construct a fixed web-search context for all open-weight LLMs. This context is prepended to the prompt in the form: \texttt{``Here is some web search context that may help you: \{scraped contents from cited URLs\}.''}

\paragraph{Code Interpreter Implementation}
We enable Code Interpreter for GPT-4.1 and GPT-5 via the official Responses API, with tool choice set as \emph{required}.

\paragraph{Web Search + Code Interpreter Implementation}
Providing Web Search and Code Interpreter simultaneously does not guarantee that both tools are invoked during generation, even when tool choice is set as \emph{required}. To control costs and ensure consistent retrieval, we therefore reuse the web-search context collected from the Web Search-only setting and prepend it to the prompt for GPT models, while enabling Code Interpreter with tool choice set as \emph{required}.






\section{Prompts\label{app:prompts}}

\subsection{Default Prompt}

\noindent We use the following prompt template in the main experiments in Section~\ref{sec:experiments}. The placeholder \texttt{\{constraints\}} is replaced with the instance-specific constraint set at inference time.

\begin{lstlisting}[style=promptstyle]
From the time-related constraints below, determine all valid date(s) (if any) that satisfy them. Depending on the conditions, the result may be zero, one, or multiple dates. Unless otherwise stated, interpret all constraints using the Gregorian calendar.

Note: Seasons are defined as:
- Winter: December, January, February
- Spring: March, April, May
- Summer: June, July, August
- Autumn: September, October, November

The constraints are as follows:

{constraints}

Carefully review the constraints and reason step-by-step to identify all valid date(s). After thorough consideration, end your response on a new line with "MY ANSWER: " followed by the valid date(s) in the format "YYYY-MM-DD". If there are multiple valid dates, list them separated by commas. If no valid date exists, respond with "MY ANSWER: None".
\end{lstlisting}

\subsection{Prompt with Step-by-step Reasoning Instruction}

The following prompt is used in the follow-up experiment in Section~\ref{sec:analysis}, which builds on the default prompt by inserting detailed step-by-step reasoning instruction.

\begin{lstlisting}[style=promptstyle]
From the time-related constraints below, determine all valid date(s) (if any) that satisfy them.
Depending on the conditions, the result may be zero, one, or multiple dates.
Unless otherwise stated, interpret all constraints using the Gregorian calendar.

Carefully follow these step-by-step reasoning instructions:

1) Read and restate the goal
   - You must find ALL Gregorian calendar date(s) that satisfy EVERY constraint.
   - The answer can be none, one, or many dates.

2) Normalize definitions you must use
   - Seasons:
     * Winter = Dec, Jan, Feb
     * Spring = Mar, Apr, May
     * Summer = Jun, Jul, Aug
     * Autumn = Sep, Oct, Nov
   - If constraints mention weekday, leap year, end of month, etc., use standard Gregorian rules.

3) Extract constraints into a checklist
   - Parse the text and list each constraint as a separate item.
   - If any constraint contains an OR, split into separate branches.
   - If constraints contain ranges (e.g., "between 2010 and 2015"), record inclusivity/exclusivity exactly as stated.

4) Determine the search space (candidate window)
   - Identify the tightest restriction on:
     * Year(s) (specific year, range, century, decade)
     * Month(s) (specific month(s), season, quarter)
     * Day(s) (day number, end of month, first Monday, etc.)
   - Use these to form an initial set of candidate dates or a candidate-generating rule.

5) Convert each constraint into a filter function
   For each constraint, translate it into a precise test such as:
   - Year filter (e.g., year = 2024; year in [1990, 1999])
   - Month/season filter (e.g., month in {{6,7,8}})
   - Day-of-month filter (e.g., day = 1; day >= 28)
   - Weekday filter (e.g., weekday = Monday)
   - Relative-date relations (e.g., "the day after X", "two weeks before Y")
   - Calendar properties:
     * Leap year rule: leap if divisible by 4, except centuries not divisible by 400.
     * Month lengths: 30/31 days; February 28 or 29 depending on leap year.

6) Generate candidates in the most restrictive way
   - Prefer generating candidates from the most specific constraints first:
     * If a specific date is implied, compute it directly.
     * If a specific year + season is given, iterate only those months.
     * If a weekday pattern exists ("first Monday of May"), generate those directly instead of brute-forcing all days.

7) Apply constraints systematically
   - For each candidate date, check constraints one by one.
   - Eliminate candidates immediately when one constraint fails.
   - Keep a running list of remaining candidates.

8) Handle OR and branching logic correctly
   - If constraints are like: (A OR B) AND C:
     * Compute candidates satisfying A and C
     * Compute candidates satisfying B and C
     * Take the union of those results (deduplicate).

9) Check for hidden contradictions early
   - Before enumerating too much, look for impossibilities, e.g.:
     * Month is February but day is 30
     * Season says Winter but month says July
     * Date required to be both Monday and Tuesday
     * Before 2000 AND after 2010
   - If a contradiction is found, conclude no solution.

10) Ensure completeness (don't miss valid dates)
   - Confirm you explored all years/months allowed by the constraints.
   - If you used branching, ensure all branches were evaluated.

11) Sort and format final answers
   - Sort valid dates chronologically.
   - Output dates in YYYY-MM-DD format.
   - If multiple, separate by commas and a space.

The constraints are as follows:

{constraints}

After thorough consideration, end your response on a new line with "MY ANSWER: "
followed by the valid date(s) in the format "YYYY-MM-DD".
If there are multiple valid dates, list them separated by commas.
If no valid date exists, respond with "MY ANSWER: None".
\end{lstlisting}

\end{document}